\documentclass[11pt]{article}
\usepackage[preprint]{acl}
\usepackage{times}
\usepackage{latexsym}
\usepackage[T1]{fontenc}
\usepackage[utf8]{inputenc}
\usepackage{microtype}
\usepackage{inconsolata}
\usepackage{graphicx}
\usepackage{amsmath,amssymb,mathtools,amsthm}
\usepackage{booktabs,multirow,array}
\usepackage{subcaption}
\usepackage{xspace,xcolor}
\usepackage{placeins}
\usepackage{tcolorbox}
\tcbuselibrary{breakable}

\newcommand{\mname}{\textsc{NHNM}\xspace}
\newcommand{\dname}{\textsc{NHNM-MARCO}\xspace}

\definecolor{promptgray}{RGB}{240,240,240}
\newtcolorbox{promptexample}{breakable,colback=promptgray,
  colframe=promptgray!50,boxrule=0.5pt,left=0.5em,right=0.5em,
  top=0.5em,bottom=0.5em}
\theoremstyle{plain}

\theoremstyle{definition}

\theoremstyle{remark}

\title{Natural Hard-Negative Mining from Positives for Instruction-Following Retrieval}
\author{Chaolang Li \quad Jingyu Li \quad Zhaocheng Du \quad Xuyang Xie \quad Gang Wang \quad Pengwen Dai}
\hypersetup{pdfauthor={Chaolang Li, Jingyu Li, Zhaocheng Du, Xuyang Xie, Gang Wang, Pengwen Dai},pdftitle={Natural Hard-Negative Mining from Positives for Instruction-Following Retrieval}}

\begin{document}
\maketitle
\linespread{0.98}\selectfont
\begin{abstract}
Instruction-following information retrieval (IFIR) requires retrievers to
rank documents by both a query and its instance-specific instructions, yet
instruction-conditioned contrastive supervision remains scarce. Existing
synthetic methods typically fix a positive passage and generate additional
negatives, overlooking the hard-negative potential of the original positive.
We propose \textbf{\textit{Natural Hard-Negative Mining} (\mname)}, which
promotes convertible negatives under query-compatible instructions and
repurposes original positives as natural instruction hard negatives. Its
automated \textit{Candidate Cleaning}, \textit{Reversal Construction}, and
\textit{Geometry Filtering} pipeline produces compact, high-contrast
supervision aligned with the pretrained retrieval space. Applied to MS MARCO,
\mname yields \dname, an IFIR training set of nearly 7K instances. On
InstructIR and FollowIR, models with 568M to 4B parameters fine-tuned on
\dname achieve strong instruction-following performance and match or surpass
substantially larger retrievers in several settings. Controlled analyses and
ablations attribute these gains to natural hard-negative supervision rather
than model scale or training-data volume.
\end{abstract}

\section{Introduction}

Instruction-following information retrieval (IFIR) requires retrievers to
satisfy both a query and user instructions about aspects, scope, evidence, or
exclusions. It supports personalized search~\cite{2025/SPARK,bok/2022/personalized},
task-oriented information seeking~\cite{feng/2023/task-oriented-dialogues,asai/2023/task-aware},
and retrieval-augmented generation~\cite{lewis/2020/RAG}.

\begin{figure}[t]
    \centering
    \includegraphics[width=\linewidth,trim=0 4bp 0 9bp,clip]{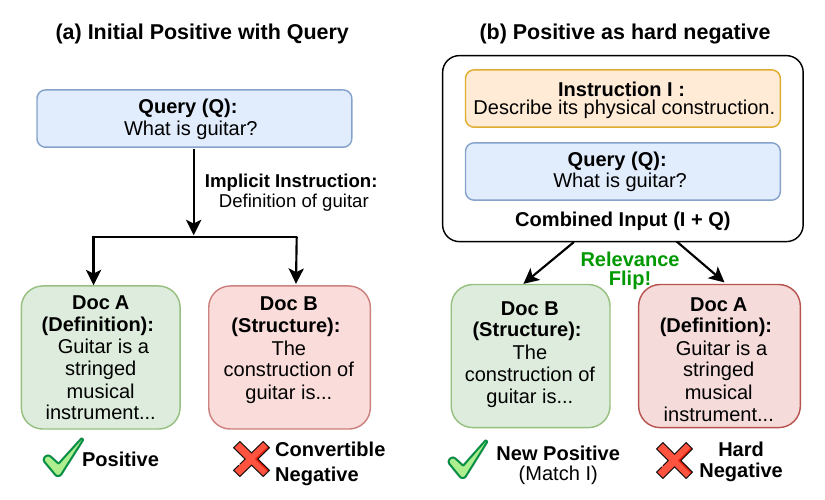}
    \caption{Illustration of instruction-conditioned natural hard-negative mining.}
    \label{fig:introduction}
\end{figure}

Despite recent progress~\cite{su/2023/Instructor,asai/2023/task-aware,muennighoff/2024/GRIT},
retrievers often remain dominated by query-document similarity and fail to
adjust rankings to instance-specific instructions~\cite{Hanseok/2024/InstructIR}.
Instruction-aware supervision is therefore needed~\cite{weller/2025/FollowIR},
but contrastive pairs whose relevance changes with the instruction are scarce
and costly to annotate. Simply adding instructions to conventional tuples with
fixed relevance labels lets models ignore them; effective supervision must
expose \emph{instruction-conditioned relevance changes}.

Existing methods create such changes around a fixed positive. Promptriever
preserves it under a generated instruction and synthesizes query-relevant,
instruction-violating negatives~\cite{weller/2025/Promptriever}; InF-IR
generates hard negatives by perturbing the instruction or query of a positive
triplet~\cite{zhuang/2025/InFIR}. Both retain a designated positive and build
negatives around it, overlooking the original positive's hard-negative
potential. Passage-level synthesis also requires a long, coherent passage that
remains query-relevant while violating the instruction, increasing exposure to
generation bias and hallucination.

Figure~\ref{fig:introduction} illustrates our key observation: an original
positive can become a \emph{natural hard negative} under an instruction whose
added criterion it violates, despite remaining query-relevant. We identify
\emph{convertible negatives} (CNs), existing negatives that become positives
under that instruction, to form the contrast. Generating a short instruction
from the semantic difference between the passages replaces controlled negative
passage synthesis and reduces generation artifacts while exploiting the
original positive's inherent hardness.

We therefore introduce \textbf{\emph{Natural Hard-Negative Mining (\mname)}}.
Its automated pipeline (1) cleans candidates and removes false negatives,
(2) selects a CN, generates a minimal instruction that promotes it, and makes
the original positive an instruction hard negative, and (3) retains instances
whose instruction-induced shift aligns with the pretrained retrieval space.
This yields precise, high-contrast supervision without synthesizing long
negative passages.

Applying \mname to MS~MARCO~\cite{bajaj/2016/ms-marco} produces the
6,716-instance \dname dataset. It improves Qwen3-Embedding-0.6B,
Qwen3-Embedding-4B, and BGE-M3; the 0.6B model is strong against much larger
retrievers on InstructIR and competitive on FollowIR.

Our contributions are threefold:
\begin{itemize}
    \item We propose \textbf{\emph{Natural Hard-Negative Mining}}, a paradigm
    that promotes convertible negatives and repurposes original positives as
    instruction hard negatives.

    \item We develop a fully automated pipeline with three simple stages:
    \textit{Candidate Cleaning}, \textit{Reversal Construction}, and
    \textit{Geometry Filtering}, producing high-contrast supervision without
    synthesizing long negative passages.

    \item We construct the compact, 6,716-instance \dname training set and show
    that it improves multiple encoder backbones; a 0.6B model is strong against
    much larger retrievers on InstructIR and competitive on FollowIR.
\end{itemize}

\section{Related Work}

\paragraph{Instruction-Aware Retrieval and Evaluation.}
Instruction-aware retrieval extends semantic matching by conditioning
relevance on a natural-language description of the user's task or preference.
Early methods such as Instructor~\cite{su/2023/Instructor} and
TART~\cite{asai/2023/task-aware} prepend task-level instructions to queries
during representation learning, while GRIT~\cite{muennighoff/2024/GRIT}
jointly learns generative and embedding capabilities under diverse
instructions. These approaches demonstrate that instructions can improve
transfer across retrieval tasks, but task-level descriptions do not
necessarily require a retriever to change its ranking for an individual
query. InstructIR~\cite{Hanseok/2024/InstructIR} therefore evaluates
instance-level instructions that express user-specific intents and
preferences. FollowIR~\cite{weller/2025/FollowIR} uses detailed TREC
narratives and controlled instruction perturbations to measure whether
rankings change in the required direction. Together, these benchmarks show
that strong conventional relevance does not imply instruction sensitivity and
motivate training supervision in which document relevance changes with the
instruction.

\paragraph{Instruction-Conditioned Data Synthesis.}
Recent work constructs such supervision automatically rather than relying on
costly relevance annotation. Promptriever~\cite{weller/2025/Promptriever}
keeps an original positive passage relevant under a generated instruction and
synthesizes query-relevant passages that violate that instruction, thereby
creating contrastive examples in which the instruction is necessary.
InF-IR~\cite{zhuang/2025/InFIR} begins with a positive
instruction--query--passage triplet and generates hard negatives by
perturbing the instruction or query. Both methods establish the value of
instruction-conditioned hard negatives, but follow a positive-centered
construction: the designated positive remains fixed, while additional
negative passages are generated around it. This construction leaves the
potential role reversal of existing retrieval candidates unexplored and
requires the LLM to synthesize long passages that are simultaneously
query-relevant and instruction-violating.

Our work investigates the reverse direction. Rather than generating a passage
around a fixed positive, we identify a \emph{convertible negative} among
existing retrieval candidates and generate a compatible instruction that
promotes it. Under the same instruction, the original query-positive passage
$d^+$ remains topically relevant but violates the added criterion, becoming a
natural instruction hard negative. This label reversal produces a
high-contrast pair of corpus-grounded documents for which the instruction,
rather than topical similarity alone, determines the preferred ranking.

\begin{figure*}[t]
    \centering
    \includegraphics[width=0.98\textwidth,trim=0 4bp 0 3bp,clip]{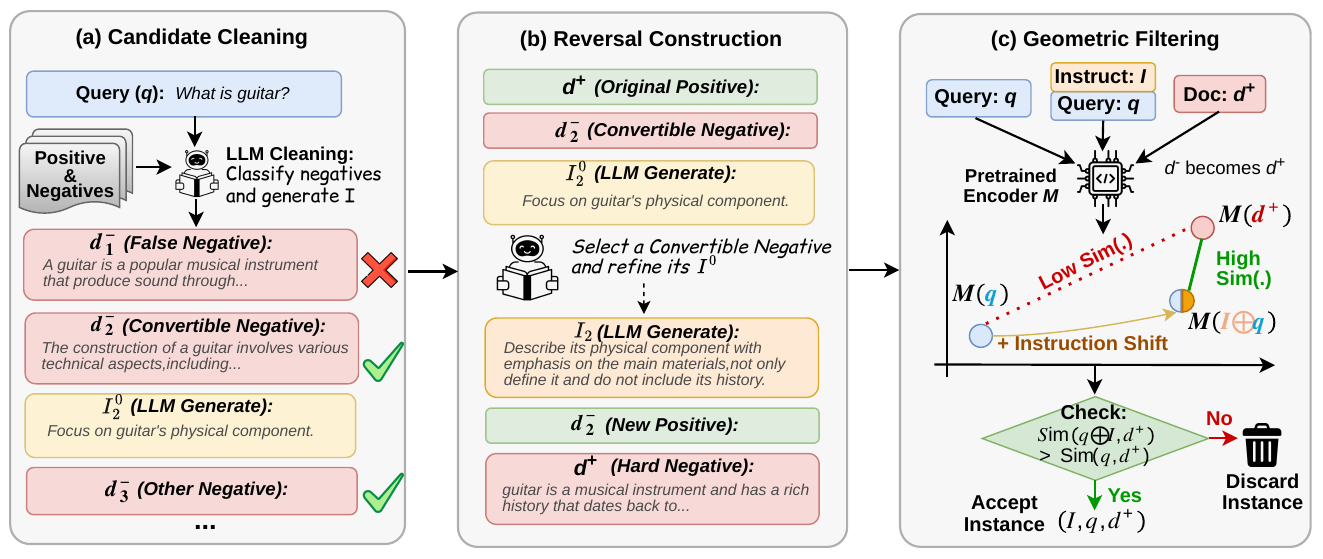}
    \caption{Overview of the \mname pipeline for mining natural hard negatives from query-positive-negative tuples.}
    \label{fig:pipeline}
\end{figure*}

\section{Preliminaries}
\label{sec:preliminaries}

\paragraph{Instruction-Following Retrieval.}
Given an instruction $I$ and a query $q$, instruction-following information
retrieval (IFIR) ranks a candidate document list according to both the
information need in $q$ and the relevance criteria in $I$. We denote the
training candidate list as
$\mathcal{D}_{I,q}=\{d^+\}\cup\{d_i^-\}_{i=1}^{K}$, where $d^+$ is the
instruction-conditioned positive document and $\{d_i^-\}_{i=1}^{K}$ are
negative documents. For a dual-encoder retriever $f_\theta$, the score of a
candidate document $d\in\mathcal{D}_{I,q}$ is
\begin{equation}
    s_\theta(I,q,d)
    =
    \operatorname{sim}\!\bigl(
        f_\theta(I\oplus q), f_\theta(d)
    \bigr),
    \label{eq:ifir_score}
\end{equation}
where $\oplus$ denotes concatenation and $\operatorname{sim}$ is cosine
similarity. We optimize the standard InfoNCE objective~\cite{oord2019infonce}:
\begin{equation}
    \mathcal{L}
    =
    -\log\!
    \frac{\exp(s_\theta(I,q,d^+)/\tau)}
    {\sum_{d\in\mathcal{D}_{I,q}}\exp(s_\theta(I,q,d)/\tau)},
    \label{eq:infonce}
\end{equation}
where $\tau$ is the temperature.

\paragraph{Instruction-Conditioned Hard Negatives.}
Instruction-conditioned mining asks whether a negative document in the
candidate list can become the preferred document under a suitable instruction.
For a negative $d_j^-\in\mathcal{D}_{I,q}$, we call $d_j^-$
\emph{convertible} if a query-compatible instruction $I_j$ can make it satisfy
the intended criterion better than the original positive $d^+$. Under the same
instruction, $d^+$ remains topically relevant to the query but violates the
added criterion, so it becomes a natural instruction hard negative.

\section{Method and Data Curation Pipeline}
\label{sec:method}

\textbf{Negative Retrieval.}
Before applying the pipeline, we retrieve candidate negatives to search for
potential convertible negatives (CNs). For each query $q$, we use BM25 to
retrieve the top-$M$ documents and exclude ground-truth positives. The
resulting documents are topically related to $q$ but do not satisfy the
original retrieval intent, often reflecting partial relevance, alternative
aspects, or complementary information. This retrieval step provides a
semantically close candidate pool in which some negatives may be promoted by
query-compatible instructions.

Figure~\ref{fig:pipeline} presents our fully automated three-stage pipeline:
\textit{Candidate Cleaning} (\S\ref{ssec:cleaning}), \textit{Reversal Construction}
(\S\ref{ssec:reversal}), and \textit{Geometry Filtering}
(\S\ref{ssec:filtering}). The first two stages determine whether a valid
reversal exists and express it as an instruction; the last stage selects
samples whose instruction-induced shifts preserve the pretrained retrieval
space. The appendix provides the prompts.

\subsection{Stage I: Candidate Cleaning}
\label{ssec:cleaning}

Turning the original positive $d^+$ into a natural hard negative first
requires finding a retrieved negative that can be validly promoted. Given
$\mathcal{D}_q=\{d^+\}\cup\mathcal{D}_q^-$, where
$\mathcal{D}_q^-=\{d_i^-\}_{i=1}^{K}$, the retrieved negatives may be noisy,
may already answer $q$, or may be topically related but incompatible with any
valid refinement. Generating instructions from them indiscriminately can force
the LLM to produce constraints that are incompatible with, or even contradict,
the original query, thereby introducing noisy instruction-following supervision.

We therefore apply hybrid rule-LLM cleaning, as illustrated in
Figure~\ref{fig:pipeline}(a). Lightweight rules first remove malformed,
truncated, and near-duplicate passages. The LLM then jointly examines
$(q,d^+,\mathcal{D}_q^-)$, verifies the validity of $d^+$, and assigns every
remaining $d_i^-$ to exactly one category: a \textbf{False Negative (FN)}
already satisfies $q$; a \textbf{Convertible Negative (CN)} does not satisfy
$q$ but can become preferable to $d^+$ under a compatible refinement; and an
\textbf{Other Negative (ON)} satisfies neither condition.

For each CN, the LLM additionally proposes an initial instruction $I_i^0$ as
a feasibility check: it must preserve the entity, intent, and answer type of
$q$, introduce a condition supported by $d_i^-$, and keep $d^+$ relevant but
noncompliant. We retain only tuples with a valid $d^+$ and at least one
feasible CN. The stage removes FNs and outputs the cleaned CN\hspace{0pt}--ON pool
$\widetilde{\mathcal{D}}_q^-$ together with the verified candidates
$\{(d_i^-,I_i^0)\}$ for Stage~II.

\subsection{Stage II: Reversal Construction}
\label{ssec:reversal}

The cleaning stage may identify multiple CNs, but not all of them yield equally
clear supervision. As illustrated in Figure~\ref{fig:pipeline}(b), a second
LLM compares all surviving CN--instruction candidates
$(d_i^-,I_i^{0})$ and selects the pair $(d_j^-,I_j^{0})$ that supports the
clearest instruction-conditioned contrast while preserving the query framework. It then
refines $I_j^{0}$ into a concise instruction $I_j$ containing the minimum
sufficient discriminating condition. We reject instructions that shift the
query intent, leak document-specific answers, or leave the original positive
satisfying the added criterion.

After reversal, we form a new IFIR training instance by relabeling the
selected CN $d_j^-$ as its instruction-conditioned positive $d^+$. The
original query-positive document is relabeled as the first
instruction-conditioned negative $d_0^-$:
\begin{equation}
    \mathcal{D}_{I_j,q}^{-}
    =
    \{d_0^-\}
    \cup
    \bigl(\widetilde{\mathcal{D}}_{q}^{-}
    \setminus \{d_j^-\}
    \setminus \mathcal{E}_{I_j}\bigr),
    \label{eq:negative_set}
\end{equation}
where $\mathcal{E}_{I_j}$ contains unselected CNs that also fully satisfy
$(I_j,q)$ and would otherwise become false negatives. Because $d_0^-$ was
originally annotated as relevant to $q$, it is naturally close to the query
but violates $I_j$, making it a high-value instruction hard negative. The final
training list follows the notation in Preliminaries:
$\mathcal{D}_{I_j,q}=\{d^+\}\cup\mathcal{D}_{I_j,q}^{-}$. Thus,
Equation~\ref{eq:negative_set} obtains the desired contrast without
synthesizing a negative passage: the LLM only generates the shorter instruction
that separates two real documents.

\subsection{Stage III: Geometry Filtering}
\label{ssec:filtering}

Semantic verification does not guarantee that a generated instruction produces
a stable and useful shift in the pretrained retrieval space. We therefore use
the frozen encoder before fine-tuning to remove samples whose instruction
signal is unlikely to align with the model's existing geometry. For each
generated tuple, we compute the instruction-induced gain of its relabeled
positive:
\begin{equation}
    \begin{aligned}
    \Delta^{+}(I_j,q,d^+)
    &=
    \operatorname{sim}\!\left(f_0(I_j\oplus q),f_0(d^+)\right)\\
    &\quad-
    \operatorname{sim}\!\left(f_0(q),f_0(d^+)\right),
    \end{aligned}
    \label{eq:geometric_gain}
\end{equation}
where $f_0$ is the pretrained encoder. We retain an instance when
\begin{equation}
    \Delta^{+}(I_j,q,d^+) > \gamma,
    \label{eq:geometric_filter}
\end{equation}
and use $\gamma=0$ in our main experiments. This criterion asks only whether
the frozen encoder already interprets the instruction in the intended direction; it
does \emph{not} require the model to rank $d^+$ above $d_0^-$ before
fine-tuning. Requiring the latter would remove precisely the unsolved
reversals that provide useful learning signal. Geometry filtering therefore
complements the LLM check: Stage~I--II ensures semantic validity, whereas
Stage~III keeps samples that provide instruction supervision while minimizing
unnecessary disruption to the pretrained retrieval representation space.

\subsection{\dname Construction}
\label{ssec:train_data}

We apply the pipeline to 36,306 tuples sampled from
MS MARCO-W~\cite{weller/2025/Promptriever}. Each tuple consists
of a query, one positive passage, and its provided negative passages and we use
only these query-positive-negative fields and do not use the accompanying
instruction annotations.
Rule-based cleaning retains 29,902 tuples, after which LLM verification identifies 7,527 
valid seeds containing 11,254
CN candidates. We use GPT-5.1 for both semantic verification and instruction
generation. The \textit{Reversal Construction} stage succeeds for 7,470 seeds,
and \textit{Geometry Filtering} retains 6,716 final training instances. Each instance
contains the generated instruction and original query, the selected CN as its
positive, and the original positive followed by the remaining valid passages
as negatives. 

To assess semantic quality beyond the automatic filters, we randomly sample
100 final instances for manual verification. An instance is considered valid
only if the generated instruction preserves the original query intent, makes
the selected CN preferable to the original positive, renders the original
positive a valid natural hard negative, and introduces no unsupported
constraint or answer leakage. Under these criteria, 88 instances pass all
checks, corresponding to a manual acceptance rate of 88\%.

We use \dname to fine-tune Qwen3-Embedding-0.6B and
Qwen3-Embedding-4B~\cite{qwen3-embedding}
with the objective in Equation~\ref{eq:infonce} to examine transfer across
scales, and additionally evaluate transfer to BGE-M3. Additional training
details are provided in the appendix.

\section{Experiments}

\subsection{Experimental Setup}

\paragraph{Training Setup.}
We use Qwen3-Embedding-0.6B and Qwen3-Embedding-4B~\cite{qwen3-embedding} as the main
backbones trained with \dname. The 0.6B model is fully fine-tuned for one
epoch, while the 4B model uses rank-8 LoRA ($\alpha=32$) for one epoch. To
test transfer to another encoder family, we also fine-tune BGE-M3
\cite{chen/2024/bge-m3} with \dname using the same LoRA setting.
All models are optimized with InfoNCE, an effective batch size of 32, AdamW
with a learning rate of $6\times10^{-6}$ and weight decay of 0.1. Unless stated
otherwise, ablations use the 0.6B backbone for one epoch with the same
optimization settings. All experiments are conducted on a single NVIDIA RTX
3090 GPU. The appendix lists the full hyperparameters.

\paragraph{Evaluation Benchmarks and Metrics.}
We evaluate on two instruction-following retrieval benchmarks. InstructIR
\cite{Hanseok/2024/InstructIR} uses short persona- and intent-oriented
instructions to test whether retrievers preserve relevance while following
query-specific constraints. We report Robustness@10 and nDCG@10, and add MAP 
for diagnostic ablations. FollowIR
\cite{weller/2025/FollowIR} contains longer topic descriptions and controlled
instruction perturbations over Robust04, News21, and
Core17~\cite{voorhees/2005/trec-robust04,allan/2017/trec-core17,trec-news21}.
We report MAP for Robust04 and Core17, nDCG@5 for News21, and $p$-MRR to
measure whether ranking changes follow the instruction perturbation.

\subsection{Baseline Models}
We compare against strong instruction-aware retrieval baselines from multiple
model families. On InstructIR, we include cross-encoder rankers such as
Mistral-7B-Instruct and FollowIR-7B, as well as bi-encoder retrievers including
BM25, Contriever, GTR, RepLLaMA, TART-Dual, Instructor, E5-Mistral,
Promptriever, BGE-M3, and Qwen3-Embedding. On FollowIR, we follow the original
benchmark categories and compare with instruction-following LMs, commercial
embedding APIs, retrieval models without task-specific fine-tuning, and
fine-tuned retrieval models such as TART, Instructor, InF-Embed, and \mname.
In the controlled data-paradigm comparison, ``Promptriever-style'' and
``InF-IR-style'' denote the same Qwen3-Embedding-0.6B backbone trained on
equal-sized data instantiated from the corresponding training formats.

\subsection{Main Results}
\begin{table}[t]
\centering
\fontsize{8.5}{9.5}\selectfont
\renewcommand{\arraystretch}{0.96}
\setlength{\tabcolsep}{2pt}
\begin{tabular}{@{}>{\raggedright\arraybackslash}p{0.59\columnwidth}ccc@{}}
\toprule
\textbf{Model} & \textbf{Size} & \textbf{Robust} & \textbf{nDCG} \\ \midrule
\multicolumn{4}{@{}l}{\textbf{Cross Encoder}} \\ \midrule
Mistral-7B-Instruct~\cite{jiang/2023/mistral7b} & 7B & \underline{35.30} & \underline{63.10} \\
FollowIR-7B~\cite{weller/2025/FollowIR} & 7B & \textbf{71.50} & \textbf{81.30} \\ \midrule
\multicolumn{4}{@{}l}{\textbf{Bi Encoder}} \\ \midrule
BM25~\cite{robertson/1995/BM25} & - & 26.92 & 76.01 \\
Contriever~\cite{izacard/2021/Contriever-msmarco} & 110M & 47.40 & 84.85 \\
GTR~\cite{ni/2022/GTR} & 110M & 34.06 & 73.35 \\
GTR-Large~\cite{ni/2022/GTR} & 335M & 37.56 & 75.95 \\
GTR-XL~\cite{ni/2022/GTR} & 1.5B & 38.34 & 75.20 \\
RepLLaMa~\cite{ma/2023/RepLLaMa} & 7B & 52.58 & 87.62 \\
TART-Dual~\cite{asai/2023/task-aware} & 110M & 47.46 & 84.81 \\
Instructor~\cite{su/2023/Instructor} & 110M & 23.73 & 50.44 \\
Instructor-Large~\cite{su/2023/Instructor} & 335M & 22.08 & 48.80 \\
Instructor-XL~\cite{su/2023/Instructor} & 1.5B & 21.53 & 48.63 \\
E5-Mistral~\cite{wang/2024/E5-mistral} & 7B & 55.42 & 86.33 \\
Promptriever~\cite{weller/2025/Promptriever} & 7B & 63.10 & \underline{92.10} \\
BGE-M3~\cite{chen/2024/bge-m3} & 568M & 42.29 & 80.67 \\
\quad + \dname & 568M & 45.84 & 84.31 \\
Qwen3-Embedding-4B~\cite{qwen3-embedding} & 4B & 50.63 & 84.99 \\
\quad + \dname & 4B & \underline{63.26} & 91.08 \\
Qwen3-Embedding-0.6B~\cite{qwen3-embedding} & 0.6B & 49.29 & 83.93 \\
\quad + \dname~(\textbf{\textit{ours}}) & 0.6B & \textbf{67.06} & \textbf{93.12} \\ \bottomrule
\end{tabular}
\caption{Main InstructIR results across multiple base retrievers, reported with Robustness@10 and nDCG@10.}
\label{tab:comparison_InstructIR}
\end{table}

\begin{table*}[t]
\centering
\fontsize{8.5}{9.5}\selectfont
\renewcommand{\arraystretch}{0.96}
\setlength{\tabcolsep}{2pt}
\begin{tabular}{@{}c|p{0.32\textwidth}|c|cc|cc|cc|cc@{}}
\toprule
\multirow{2}{*}{\textbf{Type}} &
  \multicolumn{1}{c|}{\multirow{2}{*}{\textbf{Model}}} &
  \multirow{2}{*}{\textbf{Size}} &
  \multicolumn{2}{c|}{\textbf{Robust04}} &
  \multicolumn{2}{c|}{\textbf{News21}} &
  \multicolumn{2}{c|}{\textbf{Core17}} &
  \multicolumn{2}{c}{\textbf{Average}} \\ \cmidrule(l){4-11}
& & & MAP & $p$-MRR & nDCG@5 & $p$-MRR & MAP & $p$-MRR & Score & $p$-MRR \\ \midrule
\multirow{6}{*}{\shortstack{Instruct\\LMs}}
& Flan-T5-Base~\cite{chung/2024/FLAN-T5} & 0.2B & 13.8 & +3.6 & 7.2 & +7.3 & 10.4 & +1.5 & 10.5 & +4.1 \\
& Llama-2-7B-Chat & 7B & 17.8 & +0.7 & 17.8 & +0.6 & 14.5 & +4.9 & 16.7 & +2.1 \\
& Flan-T5-Large~\cite{chung/2024/FLAN-T5} & 0.8B & 14.7 & +3.9 & 8.0 & \textbf{+8.9} & 11.4 & +1.3 & 11.4 & +4.7 \\
& GritLM-Reranker~\cite{muennighoff/2024/GRIT} & 7B & 23.1 & +6.9 & 20.2 & -6.7 & 18.8 & +10.4 & 20.7 & +3.5 \\
& Mistral-7B-Instruct~\cite{jiang/2023/mistral7b} & 7B & \underline{23.2} & \underline{+12.6} & \underline{27.2} & +4.8 & \underline{19.7} & \underline{+13.0} & \underline{23.4} & \underline{+10.1} \\
& FollowIR-7B~\cite{weller/2025/FollowIR} & 7B & \textbf{24.8} & \textbf{+13.7} & \textbf{29.6} & \underline{+6.3} & \textbf{20.0} & \textbf{+16.5} & \textbf{24.8} & \textbf{+12.2} \\ \midrule
\multirow{3}{*}{APIs}
& OpenAI v3 Large & - & \textbf{27.2} & -5.8 & 27.2 & -2.0 & \underline{21.6} & -0.2 & \textbf{25.3} & -2.7 \\
& Cohere v3 English & - & 22.3 & \underline{-3.6} & \underline{28.3} & \underline{+0.2} & 20.6 & \underline{+2.8} & \underline{23.7} & \underline{-0.2} \\
& Google Gecko~\cite{lee/2024/gecko} & - & \underline{23.3} & \textbf{-2.4} & \textbf{29.5} & \textbf{+3.9} & \textbf{23.2} & \textbf{+5.4} & \textbf{25.3} & \textbf{+2.3} \\ \midrule
\multirow{6}{*}{\shortstack{Retrieval\\Models\\No FT}}
& BM25~\cite{robertson/1995/BM25} & - & 12.1 & \underline{-3.1} & 19.3 & -2.1 & 8.1 & -1.1 & 13.2 & -2.1 \\
& Contriever~\cite{izacard/2021/Contriever-msmarco} & 110M & 19.7 & -6.1 & 22.9 & -2.8 & 15.3 & -2.5 & 19.3 & -3.8 \\
& E5-large-v2 & 335M & 17.4 & -4.2 & 24.3 & \textbf{+0.9} & 17.0 & \underline{+0.1} & 19.6 & \underline{-1.1} \\
& BGE-Large~\cite{xiao/2024/BGE} & 0.3B & 17.5 & -7.8 & 22.3 & \underline{+0.6} & 15.0 & \underline{+0.1} & 18.3 & -2.4 \\
& E5-Mistral~\cite{wang/2024/E5-mistral} & 7B & \underline{23.1} & -9.6 & \textbf{27.8} & -0.9 & \underline{18.3} & \underline{+0.1} & \underline{23.1} & -3.5 \\
& GritLM-7B~\cite{muennighoff/2024/GRIT} & 7B & \textbf{28.6} & \textbf{-1.7} & \underline{24.4} & -1.0 & \textbf{20.8} & \textbf{+2.6} & \textbf{24.6} & \textbf{-0.0} \\ \midrule
\multirow{5}{*}{\shortstack{Retrieval\\Models\\FT}}
& TART-Contriever~\cite{asai/2023/task-aware} & 110M & 14.3 & -9.0 & 21.8 & -3.0 & 13.3 & -3.0 & 16.5 & -5.0 \\
& Instructor-XL~\cite{su/2023/Instructor} & 1.5B & \underline{19.7} & -8.1 & \underline{26.1} & -0.9 & \underline{16.8} & +0.7 & \underline{20.9} & -2.8 \\
& InF-Embed (Llama-3.2-1B-Inst)~\cite{zhuang/2025/InFIR} & 1B & 19.1 & \textbf{+5.6} & \underline{26.1} & \textbf{+3.8} & 15.2 & +1.9 & 20.2 & \underline{+3.8} \\
& InF-Embed (Qwen2.5-3B-Inst)~\cite{zhuang/2025/InFIR} & 3B & 19.6 & \underline{+3.3} & 22.4 & +1.8 & 14.6 & \underline{+3.7} & 18.9 & +2.9 \\
& Ours (Qwen3-Embedding-0.6B) & 0.6B & \textbf{25.0} & +3.1 & \textbf{26.7} & \underline{+2.1} & \textbf{19.3} & \textbf{+7.1} & \textbf{23.7} & \textbf{+4.1} \\ \bottomrule
\end{tabular}
\caption{Main results on FollowIR across Robust04, News21, and Core17. Scores measure relevance, and $p$-MRR measures instruction sensitivity.}
\label{tab:FollowIR}
\end{table*}

Tables~\ref{tab:comparison_InstructIR} and~\ref{tab:FollowIR} compare our
\dname-trained variants with existing methods on InstructIR and FollowIR.
Published baselines are taken from the cited studies, while the Qwen and
BGE-M3 rows use our evaluation. Within each model type, the best and second-best scores
are bolded and underlined, respectively. The results support three
observations.

\paragraph{Strong results on both benchmarks.}
On InstructIR, \mname with Qwen3-Embedding-0.6B achieves the best bi-encoder
Robustness@10 and the best overall nDCG@10, outperforming Promptriever by
3.96 and 1.02 points, respectively. On FollowIR, \mname obtains the strongest
average relevance score and $p$-MRR among fine-tuned retrieval models. Its average score
of 23.7 is 2.8 points above the next fine-tuned retriever and is within 1.6
points of the best API result. Compared with the fine-tuned FollowIR-7B
cross-encoder, \mname is only 1.1 points lower in average relevance score
while using a much smaller 0.6B-parameter dual-encoder backbone.

\paragraph{Gains generalize across encoder architectures.}
The improvements are not tied to a single backbone. On InstructIR, \mname
improves Qwen3-Embedding-0.6B from 49.29 to 67.06 in Robustness@10 and from
83.93 to 93.12 in nDCG@10. Training with \dname also improves the
Qwen3-Embedding-4B backbone by 12.63 Robustness@10 points and 6.09 nDCG@10 points. With a
different encoder family, BGE-M3 improves from 42.29 to 45.84 in
Robustness@10 and from 80.67 to 84.31 in nDCG@10 after training. 
These consistent gains suggest that the mined natural hard
negatives provide model-agnostic instruction-following supervision.
\begingroup
\setlength{\parskip}{0pt}
\par
\makeatletter
\def\paragraph{\@startsection{paragraph}{4}{\z@}{1.5ex}{-1em}{\normalsize\bfseries}}
\makeatother
\paragraph{Compact data is highly efficient.}
\dname contains only 6,716 final training instances, yet it supports a compact
retriever that matches or surpasses methods built with much larger datasets or
models. Promptriever uses nearly 500K instruction-bearing instances and a 7B
retriever, while InF-IR starts from over 38K positive instruction--query--
passage triplets and constructs additional hard negatives, yielding a much
larger training corpus than ours. Despite this scale gap, \mname is stronger
on InstructIR and competitive on FollowIR. This supports the central premise:
mining corpus-grounded natural hard negatives can produce more useful
instruction-conditioned supervision than simply increasing synthetic data
volume or model size.
\endgroup

\subsection{Controlled Training-Data Paradigm Comparison}
\begin{table}[t]
\centering
\fontsize{8.5}{9.5}\selectfont
\renewcommand{\arraystretch}{0.96}
\setlength{\tabcolsep}{4pt}
\begin{tabular}{@{}p{0.52\columnwidth}cc@{}}
\toprule
\textbf{Training Data} & \textbf{Robust.} & \textbf{nDCG@10} \\ \midrule
No fine-tuning & 49.29 & 83.93 \\
InF-IR-style~\cite{zhuang/2025/InFIR} & 31.51 & 72.09 \\
Promptriever-style~\cite{weller/2025/Promptriever} & \underline{51.84} & \underline{86.35} \\
\dname (\textbf{\textit{ours}}) & \textbf{67.06} & \textbf{93.12} \\ \bottomrule
\end{tabular}
\caption{Controlled comparison of training-data paradigms.}
\label{tab:data_paradigm}
\end{table}

To compare training-data paradigms fairly, we remove the confounding effects of
model capacity, data volume, and optimization budget. All rows in
Table~\ref{tab:data_paradigm} use the same Qwen3-Embedding-0.6B backbone, the
same number of training instances, and the same one-epoch fine-tuning setup.
Only the construction paradigm of the training data is changed. For
Promptriever-style data, we use its instruction-bearing tuples, retaining the
original positive and the accompanying generated negatives. For InF-IR-style
data, we use its positive instruction--query--document tuples and the provided
instruction- or query-perturbed hard-negative tuples. We randomly sample each
source to the same size as
\dname and evaluate all models with the same InstructIR protocol.

\begin{figure}[t]
    \centering
    \includegraphics[width=\linewidth]{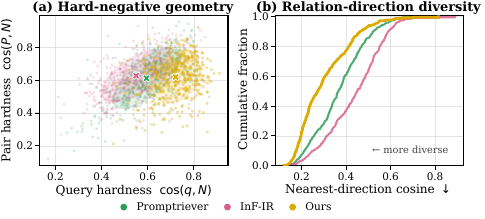}
    \caption{Geometric comparison of negative-construction paradigms.}
    \label{fig:natural_negative_geometry}
\end{figure}

Table~\ref{tab:data_paradigm} shows that Promptriever-style training provides a
modest improvement over the unfine-tuned Qwen baseline, whereas InF-IR-style
training degrades both instruction robustness and average ranking quality
under this controlled transfer. In contrast, \dname achieves the strongest
result on both metrics. Since the backbone and training budget are held fixed,
the gap reflects the supervision induced by the data paradigm itself rather
than a larger model or more training examples. Positive-centered synthetic
negatives may provide useful supervision, but they need not form a consistently
hard, corpus-grounded ordering contrast for the retriever. \mname
instead promotes a naturally retrieved convertible negative and reuses the
original query-relevant positive as an instruction-violating hard negative. The
instruction must therefore determine which of two topically compatible
documents should rank first, yielding stronger instruction-conditioned
supervision. This behavioral result is consistent with the harder and less
directionally redundant relations observed in
Figure~\ref{fig:natural_negative_geometry}.

\subsection{Ablation Study}
\begin{table}[t]
\centering
\fontsize{8.5}{9.5}\selectfont
\renewcommand{\arraystretch}{0.96}
\setlength{\tabcolsep}{2pt}
\begin{tabular}{@{}>{\raggedright\arraybackslash}p{0.27\columnwidth}|ccc|cc@{}}
\toprule
\multirow{2}{*}{\textbf{Variant}} &
  \multicolumn{3}{c|}{\textbf{InstructIR}} &
  \multicolumn{2}{c}{\textbf{FollowIR}} \\ \cmidrule(lr){2-4}\cmidrule(l){5-6}
& \shortstack{nDCG\\@10} & Robust & MAP & Score & $p$-MRR \\ \midrule
No fine-tuning & 83.93 & 49.29 & 78.79 & 22.8 & -0.2 \\
\midrule
No Cleaning & 90.62 & 60.93 & 87.64 & 22.3 & +0.2 \\
No Geometry Filter & 92.27 & 65.25 & 89.79 & 23.2 & +2.3 \\
No Original Positive & 90.31 & 60.40 & 87.22 & 22.5 & +1.8 \\
\mname (\textbf{\textit{ours}}) & \textbf{93.12} & \textbf{67.06} & \textbf{90.95} & \textbf{23.7} & \textbf{+4.1} \\ \bottomrule
\end{tabular}
\caption{Ablation results on InstructIR and FollowIR.}
\label{tab:ablation}
\end{table}

Table~\ref{tab:ablation} validates the contribution of each stage in the
\mname pipeline. The baseline is the
original Qwen3-Embedding-0.6B model without fine-tuning. ``No Cleaning''
removes candidate cleaning, ``No Geometry Filter'' removes geometry filtering,
and ``No Original Positive'' removes the original query-positive document from
the instruction-conditioned negative set.

\paragraph{Candidate cleaning.}
Removing the cleaning stage substantially reduces all InstructIR metrics and
also lowers the FollowIR relevance score from 23.7 to 22.3. This
confirms that not every annotated negative is convertible: false negatives,
corrupted passages, and topic-related but instruction-incompatible negatives
introduce noisy candidate promotions. The \textit{Candidate Cleaning} stage is
therefore necessary to ensure that the generated instruction expresses a valid
refinement rather than an artificial redefinition of the query.

\paragraph{Geometry Filtering.}
Without \textit{Geometry Filtering}, the model still benefits from
the mined supervision, improving the baseline from 49.29 to 65.25 in
Robustness@10 and from -0.2 to +2.3 in FollowIR $p$-MRR. Thus, filtering is
not what creates the training signal. However, the full pipeline further
improves Robustness@10 to 67.06, nDCG@10 to 93.12, and FollowIR $p$-MRR to
+4.1. This suggests that \textit{Geometry Filtering} favors samples whose
instruction-induced shift is compatible with the pretrained retrieval space,
helping reduce disruption to the model's original retrieval geometry while
improving instruction following.

\paragraph{Original Positive as a Natural Hard Negative.}
Removing the original positive from the instruction-conditioned negative set
weakens both InstructIR metrics and FollowIR instruction sensitivity: compared
with the full pipeline, Robustness@10 drops from 67.06 to 60.40 and
FollowIR $p$-MRR drops from +4.1 to +1.8. This is central to the proposed
\mname paradigm. The original positive is not an arbitrary negative; it remains
strongly query-relevant but violates the generated instruction. It therefore
acts as a natural hard negative that forces the retriever to distinguish
ordinary query-document matching from instruction-conditioned relevance.
Meanwhile, the FollowIR relevance score decreases more modestly, from 23.7 to
22.5, showing that the principal loss is in instruction sensitivity rather than
standard relevance effectiveness.

\begin{figure}[t]
    \centering
    \begin{subfigure}[b]{0.49\linewidth}
        \centering
        \includegraphics[width=\linewidth,trim=0 6bp 0 1bp,clip]{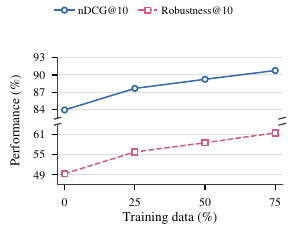}
        \caption{Results on InstructIR.}
    \end{subfigure}
    \hfill
    \begin{subfigure}[b]{0.49\linewidth}
        \centering
        \includegraphics[width=\linewidth,trim=0 6bp 0 1bp,clip]{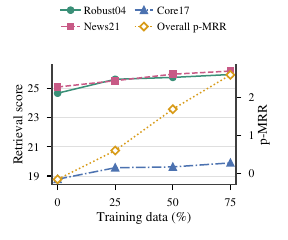}
        \caption{Results on FollowIR.}
    \end{subfigure}
    \caption{Training data size sensitivity on InstructIR and FollowIR.}
    \label{fig:data_size}
\end{figure}

\subsection{Training Data Composition Analysis}

Table~\ref{tab:composition} analyzes which training signal is responsible for
the observed gains on InstructIR. We compare three training signals against the
Qwen3-Embedding-0.6B baseline without fine-tuning: standard MS MARCO
query-positive-negative triples, instruction data generated from positive
documents, and the proposed \dname data. Fine-tuning directly on limited
MS MARCO QPN triples hurts all metrics compared with no fine-tuning,
suggesting that ordinary retrieval supervision can disturb the embedding
model's pretrained representation space without teaching instruction-
conditioned relevance. Positive-based instruction data slightly improves
nDCG@10 and MAP but not Robustness@10, indicating that instructions built
around already relevant positives provide limited contrastive semantics.

\begin{table}[t]
\centering
\fontsize{8.5}{9.5}\selectfont
\renewcommand{\arraystretch}{0.96}
\setlength{\tabcolsep}{3pt}
\begin{tabular}{@{}l|ccc@{}}
\toprule
\textbf{Training Signal} & \textbf{nDCG@10} & \textbf{Robust.} & \textbf{MAP} \\ \midrule
No fine-tuning & 83.93 & 49.29 & 78.79 \\
MS MARCO QPN & 81.80 & 45.40 & 76.12 \\
Positive Instruction & \underline{84.53} & \underline{49.07} & \underline{79.70} \\
\dname & \textbf{93.12} & \textbf{67.06} & \textbf{90.95} \\ \bottomrule
\end{tabular}
\caption{Training data composition analysis on InstructIR.}
\label{tab:composition}
\end{table}

\dname performs best across all reported metrics. Unlike the two alternatives,
it promotes a naturally retrieved convertible negative while using the
original positive as an instruction-violating natural hard negative. The model
therefore has to learn which of two query-compatible documents should be
preferred under the instruction, making the instruction indispensable rather
than auxiliary.

\subsection{Natural Hard-Negative Geometry Analysis}
Figure~\ref{fig:natural_negative_geometry} contrasts \mname with the generated
negatives of Promptriever and InF-IR using equal-sized samples encoded by a
frozen Qwen3-Embedding-0.6B model. In panel (a), crosses denote method means
and translucent ellipses summarize dispersion. \mname shifts toward
greater query--negative similarity while preserving comparable
positive--negative similarity. The original positive therefore remains a
query-relevant hard negative rather than becoming easy through topical
mismatch, supporting the motivation in Figure~\ref{fig:introduction}. Panel
(b) measures diversity through the empirical cumulative distribution of
nearest-neighbor absolute cosine between normalized positive--negative
relation directions. The leftward \mname curve indicates fewer
near-duplicate directions and hence more diverse contrastive relations.
Thus, reversal yields harder and less redundant, corpus-grounded contrasts
without synthesizing long negative passages, directly addressing the
limitations discussed in the Introduction.

\subsection{Training Data Size Sensitivity}

Figure~\ref{fig:data_size} analyzes how performance changes with the amount of
training data. We randomly sample nested subsets of \dname and fine-tune each
model for one epoch; zero percent denotes the base model without fine-tuning.
Panel (a) uses a broken shared $y$-axis to show the two metric ranges clearly.
Using only 25\% of the data already improves both benchmarks over the base
model. Performance then increases consistently from 25\% to 75\%: both
InstructIR metrics improve, while all three FollowIR collection scores and the
aggregate $p$-MRR follow the same upward trend. These results show that the
proposed construction pipeline produces effective training instances whose
benefits continue to accumulate with data size, demonstrating both the quality
and data efficiency of \dname, together with a positive scaling trend.

\FloatBarrier
\section{Conclusion}
We introduced \mname, which turns existing query-positive-negative tuples into
instruction-conditioned ordering contrasts. By promoting a convertible
negative under a concise instruction and reusing the original positive as a
topically relevant, instruction-violating hard negative, it avoids passage
synthesis and makes the instruction necessary for ranking.

Applied to MS MARCO, the pipeline produces the compact, 6,716-instance
\dname training set. Across InstructIR and FollowIR, it improves encoder
backbones from 568M to 4B parameters. Controlled comparisons and ablations
attribute these gains to the data paradigm and identify the original positive
as the central source of instruction-conditioned contrast. This validates the
proposed paradigm.

\section*{Limitations}
\label{sec:limitations}
Due to computational cost, we have not evaluated substantially larger training
corpora or backbones. We leave such large-scale data construction and
evaluation to future work.

\bibliography{references}
\clearpage
\appendix
\section{Training Details}
\label{appendix:training}

\paragraph{Qwen3-Embedding training.}
Before geometry filtering, \dname contains the same 7,470 instances produced
by \textit{Reversal Construction} for both Qwen backbones. Applying
Equation~\ref{eq:geometric_filter} with the frozen Qwen3-Embedding-0.6B model
removes 754 instances and retains 6,716, whereas filtering the same pool with
the frozen Qwen3-Embedding-4B model removes 815 and retains 6,655. We fully
fine-tune the 0.6B model and use LoRA ($r=8$, $\alpha=32$) for the 4B model.
Both reported results use the checkpoint after one epoch.

\paragraph{BGE-M3 training.}
For BGE-M3, we return to the same 7,470 pre-filtering \dname instances and
recompute Equation~\ref{eq:geometric_filter} with the frozen, original
\texttt{BAAI/bge-m3} checkpoint rather than reusing Qwen scores. With
$\gamma=0$, BGE-M3 removes 116 instances and retains 7,354. We optimize the
FlagEmbedding encoder objective with a passage group size of six and apply
LoRA to all linear attention projections with $r=8$, $\alpha=32$, and zero
dropout. The adapter is merged into the base checkpoint before evaluation;
retrieval uses normalized dense embeddings without lexical or sparse BGE-M3
scores. We report the checkpoint after one epoch.

Table~\ref{tab:fine_tune_config} summarizes the corresponding optimization
configurations.

\begin{table*}[t]
\centering
\small
\setlength{\tabcolsep}{2pt}
\renewcommand{\arraystretch}{1.02}
\begin{tabular}{@{}lccc@{}}
    \toprule
    \textbf{Parameter} & \textbf{Qwen-0.6B} & \textbf{Qwen-4B} & \textbf{BGE-M3} \\
    \midrule
    Base model & Qwen3-Emb.-0.6B & Qwen3-Emb.-4B & BGE-M3 \\
    Geometry encoder & Qwen-0.6B & Qwen-4B & BGE-M3 \\
    Retained instances & 6,716 & 6,655 & 7,354 \\
    Tuning & Full & LoRA ($r=8$) & LoRA ($r=8$) \\
    Epochs & 1 & 1 & 1 \\
    Batch (micro $\times$ accum.) & $8\times4$ & $1\times32$ & $8\times4$ \\
    Learning rate / decay & $6\!\times\!10^{-6}$ / 0.1 & $6\!\times\!10^{-6}$ / 0.1 & $5\!\times\!10^{-6}$ / 0.01 \\
    Optimizer / schedule & AdamW / cosine & AdamW / cosine & AdamW / linear \\
    Warmup ratio & 0.05 & 0.05 & 0.1 \\
    Objective & InfoNCE & InfoNCE & InfoNCE \\
    Pooling / norm. & Last-token / L2 & Last-token / L2 & CLS / L2 \\
    Max. query/passage length & 32,768 / 32,768 & 32,768 / 32,768 & 128 / 256 \\
    Precision & bfloat16 & bfloat16 & bfloat16 \\
    Seed & 42 & 42 & 42 \\
    \bottomrule
\end{tabular}
\caption{Fine-tuning configurations for the main results.}
\label{tab:fine_tune_config}
\end{table*}

All runs use one RTX 3090 with gradient checkpointing and maximum gradient norm
1.0.

\section{More Experiments on FollowIR}
\label{appendix:qwen4b_followir}

\begin{table*}[t]
\centering
\normalsize
\setlength{\tabcolsep}{3.1pt}
\renewcommand{\arraystretch}{1.06}
\begin{tabular}{@{}llcccc@{}}
\toprule
\textbf{Method} & \textbf{Metric} & \textbf{Robust04} & \textbf{News21} & \textbf{Core17} & \textbf{Avg.} \\
\midrule
\multirow{2}{*}{Baseline}
& Score   & 26.5 & 26.1 & 21.7 & 24.8 \\
& $p$-MRR & +3.5 & +1.2 & +4.4 & +3.0 \\
\midrule
\multirow{2}{*}{$+\dname$}
& Score   & \textbf{27.1} & \textbf{26.7} & \textbf{22.1} & \textbf{25.3} \\
& $p$-MRR & \textbf{+6.9} & \textbf{+5.8} & \textbf{+8.8} & \textbf{+7.2} \\
\bottomrule
\end{tabular}
\caption{Qwen3-Embedding-4B results on FollowIR. Score denotes MAP for
Robust04 and Core17 and nDCG@5 for News21; Avg. is the macro
average.\label{tab:qwen4b_followir}}
\end{table*}

Beyond the 0.6B backbone used in the main experiments,
Table~\ref{tab:qwen4b_followir} reports FollowIR results with
Qwen3-Embedding-4B. Fine-tuning with \dname improves both the standard
retrieval score and $p$-MRR across all three subsets. These consistent gains
show that our supervision generalizes to a larger backbone and strengthens its
response to changed relevance instructions while preserving retrieval quality.

\section{Conventional Retrieval Retention}
\label{appendix:basic_retrieval}

Although \dname is designed to improve instruction-conditioned relevance, we
also test whether its supervision preserves useful conventional retrieval
ability. We evaluate the one-epoch Qwen3-Embedding-0.6B model on the complete
8,841,823-passage MS MARCO corpus, retrieving 1,000 passages for each of 6,980
Dev, 43 DL19, and 54 DL20 queries. Embeddings are L2-normalized and ranked by
exact dot product; the maximum query and passage lengths are 32 and 156,
respectively. Following the standard setting, we report MRR@10 on Dev and
nDCG@10 on DL19 and DL20.

\begin{table*}[t]
\centering
\centering
\normalsize
\setlength{\tabcolsep}{4.0pt}
\renewcommand{\arraystretch}{1.05}
\begin{tabular}{@{}lccc@{}}
    \toprule
    \textbf{Method} & \textbf{Dev} & \textbf{DL19} & \textbf{DL20} \\
    & \textbf{MRR@10} & \textbf{nDCG@10} & \textbf{nDCG@10} \\
    \midrule
    SEAL & 12.7 & -- & -- \\
    BM25+RM3 & 15.7 & 52.2 & 49.0 \\
    BM25 & 18.4 & 50.6 & 48.0 \\
    MINDER & 18.6 & -- & -- \\
    Qwen3-Emb.-0.6B & 18.6 & -- & -- \\
    \textbf{Ours} & \textbf{25.2} & \textbf{56.4} & \textbf{51.9} \\
    \bottomrule
\end{tabular}
\caption{Conventional retrieval performance on MS MARCO passage ranking.
Scores are percentages. Dev uses MRR@10, while TREC DL19 and DL20 use
nDCG@10.}
\label{tab:basic_retrieval}
\end{table*}

Table~\ref{tab:basic_retrieval} compares with compact lexical and generative
references reported in prior work~\cite{satouf2026quester,archish2026singlevector}, rather than presenting a
comprehensive MS MARCO leaderboard. The results show that, after fine-tuning on
the instruction-oriented \dname corpus, our model retains meaningful
conventional retrieval ability on MS MARCO Dev and the densely judged TREC DL
sets. Thus, improving instruction-conditioned relevance does not eliminate its
basic topical retrieval capability.

\section{Token Usage}
\label{appendix:token_usage}
Only \textit{Candidate Cleaning} and \textit{Reversal Construction} invoke
the LLM API. Table~\ref{tab:tokens_usage} reports both the average cost per
stage input and the total usage recorded for the main training data.
\textit{Candidate Cleaning} accounts for most of the 64.98M tokens because it
processes the larger rule-filtered pool while jointly validating $d^+$,
classifying every retrieved negative, and drafting $I_0$ for each CN.
\textit{Reversal Construction} then selects a CN and refines its $I_0$ into
the final instruction. Rule-based cleaning and \textit{Geometry Filtering}
are performed locally and incur no API-token cost.
\begin{table*}[t]
\centering
\normalsize
\setlength{\tabcolsep}{2.0pt}
\renewcommand{\arraystretch}{1.05}
\begin{tabular}{@{}lrrrrr@{}}
\toprule
& \textbf{No. of} & \multicolumn{3}{c}{\textbf{Tokens per input}} & \textbf{Total} \\
\cmidrule(lr){3-5}
\textbf{API stage} & \textbf{inputs} & \textbf{Input} & \textbf{Output} & \textbf{Total} & \textbf{tokens} \\
\midrule
Candidate Cleaning     & 29,902 & 1,401.1 & 352.8 & 1,754.0 & 52.45M \\
Reversal Construction  &  7,527 & 1,303.2 & 362.3 & 1,665.5 & 12.54M \\
\bottomrule
\end{tabular}
\caption{API token consumption for constructing the training data used in
the main experiments. Per-input values are averaged over the tuples or seeds
entering each stage. Stage totals include automatic retries and are reported
in millions.}
\label{tab:tokens_usage}
\end{table*}

\section{Prompts Used in Data Construction}
\label{appendix:prompt}

Below we reproduce the two LLM prompts used by the final pipeline. Bracketed
fields are replaced with the corresponding tuple content. The implementation
additionally validates the returned JSON schema, index coverage, and index
bounds deterministically.

\subsection{Stage I: Candidate Cleaning and CN Feasibility}

\begin{promptexample}
\normalsize\raggedright
\textbf{Goal.} Obtain high-quality seeds for instruction-following retrieval.
Read every definition and step strictly.\\[2pt]
\textbf{Definitions.}
An FN is a negative $N$ that satisfies every requirement of $Q$ as well as or
better than $P$. A CN does not directly or accurately answer $Q$, but follows
its main query direction; a compatible instruction constraint $I$ can refine
$Q$ so that $N$ becomes a better positive than $P$. An ON is neither an FN nor
a valid CN, including irrelevant passages and topically related passages for
which no compatible $I$ creates this contrast.\\[2pt]
\textbf{Step 1---Validate $P$.} Judge whether $P$ directly and accurately
answers every requirement of $Q$. If it is incomplete, indirect, or
off-intent, discard the sample and do not revise $Q$ or $P$.\\[2pt]
\textbf{Step 2---Classify every $N$.} Assign each $N$ to exactly one of FN,
CN, or ON. For every proposed CN, produce a neutral, concise draft instruction
$I_0$. Accept the CN only if $I_0$: (i) preserves the main entity, intent, and
answer type of $Q$; (ii) introduces one or two discriminating semantic
dimensions supported by $N$ without revealing its answer; and (iii) makes $N$
better than $P$, while $P$ remains relevant to $Q$ alone and loses only because
it lacks the added dimension. If no $N$ qualifies as a CN, discard the sample.\\[2pt]
\textbf{Step 3---Check and return.} Recheck every $N$ for missed FNs and every
CN/$I_0$ against all three conditions. Return every valid CN rather than
selecting or ranking candidates. FN ids, CN ids, and ON ids must classify every
$N$ exactly once. Return JSON only using the schema:\\
\texttt{\{"keep\_sample": true, "fn\_ids": [1],}
\texttt{"cn\_candidates": [\{"id": 0, "instruction": "$I_0$"\}],}
\texttt{"on\_ids": [2], "reason": "brief rationale"\}}.\\[2pt]
\textbf{Input.} $Q$: [query] \quad $P$: [original positive] \\
$N_0$: [negative 0] \quad $\ldots$ \quad $N_K$: [negative $K$]
\end{promptexample}

\subsection{Stage II: Negative-to-Positive Reversal}

\begin{promptexample}
\normalsize\raggedright
\textbf{Goal.} Select one CN and generate an instruction under which it becomes
relevant while the original positive becomes non-relevant.\\[2pt]
\textbf{Step 1---Select one CN.} Compare every candidate CN jointly with its
$I_0$ and $Q$. Select the candidate supporting the clearest objective
distinction from $P$ without changing the main subject, intent, relation, or
answer type of $Q$. Topic or keyword overlap alone is insufficient.\\[2pt]
\textbf{Step 2---Rewrite $I_0$ into $I^-$.} Use $I_0$ as a semantic draft and
repair it when needed. Write one or two concise, concrete, and unambiguous
sentences specifying what a relevant document should contain, discuss,
identify, compare, or exclude. Add one principal discriminating condition, or
at most two when necessary. The condition may concern aspect, scope, entity,
location, time, perspective, evidence type, source, audience, or format, but
must be supported by $I_0$ and the selected CN rather than an incidental
passage detail. Under $I^-+Q$, the selected CN must be relevant; $P$ must remain
relevant to $Q$ alone but become non-relevant because it fails the added
condition; all other negatives must remain non-relevant unless they satisfy
substantially the same condition. State positive inclusion criteria and add an
exclusion only when needed. Do not merely paraphrase $I_0$, rewrite $Q$, use a
vague preference, reveal the answer, copy distinctive answer-bearing details,
or mention passage ids or the generation task.\\[2pt]
\textbf{Step 3---Check and return.} Verify the reversal against $P$, the
selected CN, and every remaining $N$. Put an unselected CN in
\texttt{equivalent\_cn\_ids} only if it also fully satisfies $I^-+Q$ and would
be an equally valid positive. Reject if no natural instruction exists, if $P$
still satisfies the added condition, or if the resulting negative list is
empty. Return JSON only using the schema:\\
\texttt{\{"accept": true, "selected\_cn\_id": 0,}
\texttt{"negative\_instruction": "$I^-$",}
\texttt{"equivalent\_cn\_ids": [], "reason": "brief reasoning"\}}.\\[2pt]
\textbf{Input.} $Q$: [query] \quad $P$: [original positive] \\
candidate CN ids: [ids] \quad $N_i$ [candidate $I_0$]: [passage]
\end{promptexample}

\section{Data-Construction Cases}
\label{appendix:construction_cases}
The cases below expose both decisions made by the pipeline. During
\textit{Candidate Cleaning}, each removed FN independently satisfies the
original query, whereas a retained CN supports a compatible discriminating
condition rather than mere topic overlap. In the final instances, this
condition is absent from the original $d^+$ but satisfied by the selected CN;
the former therefore remains topically close while becoming a natural hard
negative under the refined instruction.

\begin{table*}[!t]

\centering
\fontsize{10}{12}\selectfont
\setlength{\tabcolsep}{2.8pt}
\renewcommand{\arraystretch}{1.0}
\begin{minipage}[t]{0.488\textwidth}
\textbf{Candidate Cleaning: CN identification and FN removal}\par\smallskip
\begin{tabular}{@{}p{0.21\linewidth}p{0.74\linewidth}@{}}
\toprule
\textbf{Field} & \textbf{Case and decision} \\
\midrule
\multicolumn{2}{@{}l}{\textbf{1. Healing time for broken ribs}} \\
Query & ``How long does it take broken ribs to heal?'' \\
$d^+$ & Gives the usual six-to-eight-week recovery time. \\
CN & Describes fractures still unhealed after roughly two months, non-union,
and possible surgery. \\
FN removed & Independently states that fractured-rib pain and disability can
last about six weeks. \\
Why & The FN already answers the duration question; the CN supports a
delayed-healing condition absent from the ordinary answer. \\
\midrule
\multicolumn{2}{@{}l}{\textbf{2. Purpose of the Equal Rights Amendment}} \\
Query & ``What was the purpose of the Equal Rights Amendment?'' \\
$d^+$ & States that the ERA sought constitutional equality for women. \\
CN & Discusses its role in the feminist movement and its influence on later
legal and social change. \\
FN removed & Gives the amendment text prohibiting denial of legal rights on
account of sex. \\
Why & The FN directly states the purpose; the CN supports a compatible
downstream-impact condition. \\
\midrule
\multicolumn{2}{@{}l}{\textbf{3. Causes of blood in urine}} \\
Query & ``What causes blood in urine?'' \\
$d^+$ & Lists infections, kidney stones, medications, and urinary-tract
damage as possible causes. \\
CN & Explains specifically how kidney stones can injure the kidney or ureter
lining and produce bleeding. \\
FN removed & Independently lists inflammation, infection, kidney disease,
stones, injury, cancer, and bleeding disorders. \\
Why & The FN duplicates the broad answer obligation; the CN supports a
kidney-stone-specific causal refinement. \\
\bottomrule
\end{tabular}
\end{minipage}
\hfill
\begin{minipage}[t]{0.488\textwidth}
\textbf{Final instances: CN $\rightarrow d^+$ and original $d^+\rightarrow d^-$}\par\smallskip
\begin{tabular}{@{}p{0.21\linewidth}p{0.74\linewidth}@{}}
\toprule
\textbf{Field} & \textbf{Final role reversal} \\
\midrule
\multicolumn{2}{@{}l}{\textbf{1. Persistent rib fracture}} \\
Instruction & Require the usual healing time and the response to a fracture
unhealed after about two months, including non-union assessment and surgery. \\
New $d^+$ & Gives the two-to-three-month non-union threshold and surgical
intervention for a fracture showing no healing. \\
Hard $d^-$ & The original positive gives ordinary recovery time and
pain management but not persistent non-union or surgery. \\
\midrule
\multicolumn{2}{@{}l}{\textbf{2. ERA purpose and downstream impact}} \\
Instruction & Require both the ERA's purpose and its impact on the feminist
movement or subsequent legal and social changes. \\
New $d^+$ & Connects the fight for gender equality with the ERA's influence
on later legal and social change. \\
Hard $d^-$ & The original positive states the equal-rights purpose
but not its downstream impact. \\
\midrule
\multicolumn{2}{@{}l}{\textbf{3. Critical glucose levels in newborns}} \\
Instruction & Restrict critical glucose levels to newborns and distinguish
healthy full-term from premature babies. \\
New $d^+$ & Gives separate glucose ranges for full-term and premature infants
and describes neonatal hypoglycemia. \\
Hard $d^-$ & The original positive gives a general critical threshold
but no newborn- or prematurity-specific criterion. \\
\midrule
\multicolumn{2}{@{}l}{\textbf{4. An outline as a prewriting framework}} \\
Instruction & Require an outline used before drafting an essay, with main
ideas and supporting points arranged as an organizational framework. \\
New $d^+$ & Describes a decimal outline created to organize an essay writer's
thoughts and ensure that key points are not omitted. \\
Hard $d^-$ & The original positive gives only the generic dictionary sense of
an outline as a preliminary plan or version of a written work. \\
\bottomrule
\end{tabular}
\end{minipage}
\caption{Qualitative cases from the main construction run. Left: correctly
removed FNs already satisfy the original query, whereas retained CNs support
a compatible discriminating condition. Right: examples retained after
\textit{Geometry Filtering}, where that condition promotes the CN and turns
the original positive into a topically matched natural hard negative. Passage
text is shortened without changing the relevance decisions.}
\end{table*}

These cases show that cleaning is based on answer sufficiency rather than
surface similarity. The removed FNs restate or independently fulfill the
original information need, so treating them as negatives would introduce
contradictory supervision. By contrast, each retained CN supplies a distinct
facet---such as delayed healing, downstream policy impact, or a specific
causal mechanism---that remains compatible with the original query but is not
adequately covered by the ordinary answer.

The final instances further illustrate why the original positive is a useful
natural hard negative. It remains a correct and topically close answer to the
query alone, yet fails one explicit condition concerning time, population,
impact, or document function. The promoted CN satisfies that condition without
changing the underlying entity or topic. The resulting pair is therefore more
informative than an unrelated negative: the model must resolve the
instruction-induced relevance boundary rather than rely on coarse topical
dissimilarity.

\section{Case Studies}
\label{appendix:cases}

\begin{table*}[t]
\centering
\normalsize
\setlength{\tabcolsep}{6pt}
\renewcommand{\arraystretch}{1.0}
\begin{tabular}{p{0.47\textwidth}p{0.47\textwidth}}
\toprule
\multicolumn{2}{p{0.96\textwidth}}{
\textbf{InstructIR instruction.} I am a pharmacist and need detailed
scientific literature about how dihydroquercetin interacts with common
pharmaceuticals, particularly for cardiovascular and metabolic health.\par
\textbf{Query.} what is dihydroquercetin} \\
\midrule
\textbf{Ours: top-ranked documents} & \textbf{Qwen3-Embedding-0.6B baseline} \\
\midrule
\textbf{1. Relevant.} Taxifolin is discussed in terms of drug
biotransformation, liver enzymes, and interactions with medications for
cardiovascular and metabolic disorders.
&
\textbf{1. Non-relevant.} A general scientific overview of taxifolin's
antioxidant properties, natural sources, and possible health benefits. \\

\textbf{2. Non-relevant.} A broad review of taxifolin's antioxidant effects,
natural sources, cardiovascular wellness, and glycemic control.
&
\textbf{2. Non-relevant.} A dermatology-focused passage about topical
taxifolin formulations and skin health. \\

\textbf{3. Non-relevant.} A passage about anticoagulants for heart-rhythm
disorders that does not address dihydroquercetin.
&
\textbf{3. Non-relevant.} A holistic-health passage about taxifolin's
anti-inflammatory properties and compatibility with herbal extracts. \\
\bottomrule
\end{tabular}
\caption{InstructIR case study. Ours satisfies the drug-interaction
instruction; the baseline follows the broad topic ``dihydroquercetin.''}
\label{tab:case_study_instructir}
\end{table*}

\begin{table*}[t]
\centering
\normalsize
\setlength{\tabcolsep}{6pt}
\renewcommand{\arraystretch}{1.0}
\begin{tabular}{p{0.47\textwidth}p{0.47\textwidth}}
\toprule
\multicolumn{2}{p{0.96\textwidth}}{
\textbf{FollowIR query.} Evidence that radio waves from radio towers or mobile
phones affect brain-cancer occurrence.\par
\textbf{Changed relevance criterion.} Relevant documents report experimental,
statistical, or news evidence comparing brain-cancer incidence; any document
mentioning leukemia is non-relevant.} \\
\midrule
\textbf{Ours: top-ranked documents} & \textbf{Qwen3-Embedding-0.6B baseline} \\
\midrule
\textbf{1. Relevant.} ``Disconnecting Phones and Tumors'' reports a Danish
study comparing cellphone use among people with and without brain tumors.
&
\textbf{1. Relevant.} ``Cell Phone Carcinogens?'' discusses trends in brain
tumors and possible risks from electromagnetic fields. \\

\textbf{2. Relevant.} ``Really? The Claim---Cellphones Increase the Risk of
Brain Cancer'' summarizes evidence concerning the claimed association.
&
\textbf{2. Relevant.} ``Really? The Claim---Cellphones Increase the Risk of
Brain Cancer'' summarizes evidence concerning the claimed association. \\

\textbf{3. Relevant.} ``Two Studies Report No Links to Cancer in Cell Phones'
Use'' reports controlled studies of cellphone users and brain tumors.
&
\textbf{3. Relevant.} ``Disconnecting Phones and Tumors'' reports a Danish
case-control study. \\

\textbf{4. Relevant.} ``As Cell Phone Use Explodes, Debate Over Health Risk
Grows'' reviews human and animal evidence about brain cancer.
&
\textbf{4. Relevant.} ``Two Studies Report No Links to Cancer in Cell Phones'
Use'' reports controlled studies of cellphone users and brain tumors. \\

\textbf{5. Relevant.} ``No Cellphone Surprise'' comments on the reported lack
of association between cellphone use and brain tumors.
&
\textbf{5. Non-relevant.} ``Cellular Phone Scare Discounted'' foregrounds an
elevated rate of acute myeloid \emph{leukemia}, violating the added exclusion. \\
\bottomrule
\end{tabular}
\caption{FollowIR case study under a changed relevance criterion. Ours keeps
the top results aligned with brain-cancer evidence, while the baseline still
ranks a document that directly violates the leukemia exclusion.}
\label{tab:case_study_followir}
\end{table*}

The examples compare the unmodified Qwen3-Embedding-0.6B baseline with the
one-epoch \dname model used for the main results; no case-specific model or
inference setting is used. In Table~\ref{tab:case_study_instructir}, the query
alone asks for a definition of dihydroquercetin, while the instruction narrows
the need to scientific evidence about interactions with cardiovascular and
metabolic drugs. The baseline retrieves passages about the same compound but
with mismatched facets such as skincare and herbal use. Ours instead places
the annotated drug-interaction passage first. The distinction is therefore
not improved lexical matching, but selection of the passage that satisfies the
instruction-induced facet.

Table~\ref{tab:case_study_followir} illustrates controlled relevance change.
The modified criterion explicitly excludes documents mentioning leukemia.
Although the baseline's fifth result is topically close to the query, it is
centered on acute myeloid leukemia and is non-relevant under the modified
criterion. Ours removes this violating document from the leading results while
retaining documents that report evidence about mobile-phone exposure and
brain tumors. Together, the two cases illustrate the intended effect of
natural hard-negative supervision: topic-compatible passages remain close,
but their order is determined by whether they satisfy the added relevance
condition.

\FloatBarrier
\section{Ethical considerations}
\label{sec:ethics}
This paper presents work whose goal is to advance instruction-following
information retrieval. The method relies on LLM-generated instructions and
may therefore inherit ambiguity or bias from generated data. We mitigate this
risk through rule-based checks, joint semantic verification, and geometry
filtering, and do not construct evaluation labels from generated data.

A generative AI assistant was used to support the manuscript's ACL-format
migration, diagnose LaTeX layout issues, and suggest limited wording
compression in the Abstract, Introduction, and Limitations. All resulting text
and formatting are subject to author review, and the authors remain responsible
for the paper's claims, citations, and presentation. GPT-5.1 is also used as a
component of the data-construction method, as documented in Section~4.4 and
Appendix~D.

\end{document}